\documentclass{article}

\usepackage[preprint]{neurips_2026}

\usepackage[utf8]{inputenc} 
\usepackage[T1]{fontenc}    
\usepackage{amsmath}        
\newtheorem{proposition}{Proposition}
\usepackage{amssymb}        
\usepackage{hyperref}       
\usepackage{url}            
\usepackage{booktabs}       
\usepackage{amsfonts}       
\usepackage{nicefrac}       
\usepackage{microtype}      
\usepackage{xcolor}         
\usepackage{graphicx}
\usepackage{listings}
\usepackage{placeins}

\title{AdaDPO: Self-Adaptive Direct Preference Optimization with Balanced Gradient Updates}

\author{%
  Shaolong Chen \\
  Incept Labs\\
  Houston, TX\\
  \And
  Madalina Ciobanu \\
  Incept Labs\\
  Houston, TX\\
  \AND
  Qingqing Mao$^*$ \\
  Incept Labs, Houston, TX\\
  Titan Holdings, San Francisco, CA\\
  \texttt{qmao@inceptlabs.ai} \\
  \And
  Ritankar Das \\
  Incept Labs, Houston, TX\\
  Titan Holdings, San Francisco, CA\\
}

\begin{document}

\maketitle

\begin{abstract}
Direct Preference Optimization (DPO) has become a widely adopted alternative to reinforcement learning from human feedback (RLHF) for aligning large language models with human preferences, eliminating the need for a separate reward model or reinforcement learning loop. However, recent theoretical analysis uncovers an asymmetric gradient behavior in DPO: the loss suppresses dispreferred responses substantially faster than it promotes preferred counterparts, causing the model to predominantly learn to avoid bad answers rather than to generate good ones. We propose \textbf{AdaDPO}, a Self-\textbf{Ada}ptive variant of the \textbf{D}irect \textbf{P}reference \textbf{O}ptimization algorithm that introduces per-preference-pair, stop-gradient-based coefficients derived directly from the policy model's generation probabilities, with the reference model's probabilities as an optional component. AdaDPO is constructed to enforce equality of gradient magnitudes between preferred and dispreferred probabilities; the practical implementation balances per-token gradients and applies a numerical clipping bound for stability, while retaining DPO's original hyperparameter structure. On Llama-3-8B-Instruct trained on UltraFeedback under a setup nearly identical to that of SimPO, AdaDPO consistently outperforms DPO on AlpacaEval 2: it achieves higher length-controlled win rates (LC) in 81\% of hyperparameter combinations, attains the global best LC (48.3\%) and raw win rate (46.1\%), and enlarges the LC-over-WR margin in 88\% of combinations, indicating effective mitigation of length bias. Additional analyses on KL divergence, reward margin, and reward accuracy confirm that AdaDPO rectifies the gradient imbalance and yields more efficient optimization. \textbf{Because it operates purely at the loss level, AdaDPO can be dropped into existing preference-based alignment pipelines without changing data collection or model architectures.} The method requires only a few lines of code, and the same self-adaptive principle generalizes to a broad family of pairwise contrastive preference losses including SimPO, R-DPO, IPO, CPO, and ORPO.
\end{abstract}

\section{Introduction}
\label{sec:intro}

Aligning large language models (LLMs) with human preferences has become essential for deploying capable, controllable systems. Reinforcement Learning from Human Feedback~\citep{christiano2017deep, ziegler2019fine}, typically implemented with Proximal Policy Optimization~\citep{schulman2017proximal}, established the dominant paradigm but introduces substantial complexity through its reliance on a separately trained reward model and an online RL loop. Direct Preference Optimization (DPO)~\citep{rafailov2023direct} simplified this pipeline dramatically by reparameterizing the reward function and learning the policy directly from a fixed preference dataset via a binary cross-entropy loss, achieving strong empirical performance without reinforcement learning.

Despite DPO's success, recent theoretical analysis~\citep{feng2024towards} has revealed an asymmetric gradient pathology in its loss function. Using a field-theoretic analysis of DPO's gradient vector field, \citet{feng2024towards} show that the loss applies a substantially larger gradient to dispreferred responses than to preferred ones, with the asymmetry growing as the policy becomes more confident. As a consequence, DPO-trained models predominantly learn to suppress bad answers rather than to promote good ones---an effect that becomes especially pronounced when preferred and dispreferred responses are semantically close. This is not an artifact of hyperparameter choice but a structural property of the loss: the same coefficient $\beta$ multiplies both log-ratios, leaving no mechanism to balance the per-pair gradient magnitudes. Several DPO variants address related concerns: SimPO~\citep{meng2024simpo}, R-DPO~\citep{park2024disentangling}, IPO~\citep{azar2023general}, CPO~\citep{xu2024contrastive}, and ORPO~\citep{hong2024orpo}. However, none directly addresses the per-preference-pair gradient imbalance: the asymmetry persists in any pairwise contrastive loss with a shared coefficient on both terms.

We propose AdaDPO, a self-adaptive variant of DPO that resolves this imbalance by replacing the fixed $\beta$ in DPO's implicit reward margin with per-preference-pair coefficients $\beta_w$ and $\beta_l$. Setting $\beta_l = \beta$ (preserving DPO's hyperparameter) and computing $\beta_w$ via a stop-gradient-based ratio of generation probabilities, AdaDPO is constructed to enforce equality of gradient magnitudes on preferred and dispreferred probabilities at every optimization step (the length-normalized variant used in our experiments balances gradients per-token within a numerical clipping bound). The construction is theoretically principled, requires only a few lines of code to implement, and introduces no sensitive hyperparameters beyond those already in DPO. The same self-adaptive principle extends as a drop-in modification to any pairwise contrastive preference loss (SimPO, R-DPO, IPO, CPO, ORPO).

\paragraph{Contributions.}
\textbf{(i)} We rigorously derive the condition under which a DPO-style loss produces equal gradient magnitudes on preferred and dispreferred probabilities, and prove that AdaDPO satisfies it exactly through stop-gradient operators, providing a \emph{principled} correction of the gradient imbalance (Section~\ref{sec:method}).
\textbf{(ii)} On Llama-3-8B-Instruct trained on UltraFeedback under a setup nearly identical to SimPO~\citep{meng2024simpo}, AdaDPO consistently outperforms DPO on AlpacaEval 2: higher length-controlled win rates in 81\% of hyperparameter combinations, the global best LC (48.3\%) and raw WR (46.1\%), and an enlarged LC-over-WR margin in 88\% of combinations, indicating effective mitigation of length bias (Section~\ref{sec:results}).
\textbf{(iii)} The same self-adaptive principle applies as a drop-in modification to any pairwise contrastive preference loss; we provide explicit constructions for SimPO, R-DPO, IPO, CPO, and ORPO that preserve each method's existing hyperparameters (Section~\ref{sec:generalization}), highlighting a general recipe for balancing gradients in preference optimization.

\section{Background}
\label{sec:background}

\paragraph{Direct Preference Optimization.}
We are given a preference dataset $\mathcal{D} = \{(x, y_w, y_l)\}$ of triples consisting of a prompt $x$, a preferred response $y_w$, and a dispreferred response $y_l$, together with a reference policy $\pi_{\mathrm{ref}}$. DPO~\citep{rafailov2023direct} optimizes the policy $\pi_\theta$ by minimizing
\begin{equation}
\label{eq:dpo}
\mathcal{L}_{\mathrm{DPO}}(\pi_\theta; \pi_{\mathrm{ref}}) = -\mathbb{E}_{(x,y_w,y_l) \sim \mathcal{D}} \left[ \log \sigma\!\left( \beta \log \frac{\pi_\theta(y_w \mid x)}{\pi_{\mathrm{ref}}(y_w \mid x)} - \beta \log \frac{\pi_\theta(y_l \mid x)}{\pi_{\mathrm{ref}}(y_l \mid x)} \right) \right],
\end{equation}
where $\sigma$ is the sigmoid function and $\beta > 0$ controls the deviation from $\pi_{\mathrm{ref}}$. The same coefficient $\beta$ multiplies both log-ratios---a structural choice with consequences we examine next.

\paragraph{The gradient imbalance pathology.}
\citet{feng2024towards} analyze the gradient vector field of $\mathcal{L}_{\mathrm{DPO}}$ and identify a structural asymmetry. Let $P_w \!=\! \pi_\theta(y_w \mid x)$, $P_l \!=\! \pi_\theta(y_l \mid x)$, $R_w \!=\! \pi_{\mathrm{ref}}(y_w \mid x)$, $R_l \!=\! \pi_{\mathrm{ref}}(y_l \mid x)$, and let $x_w \!=\! P_w/R_w$, $x_l \!=\! P_l/R_l$. The partial derivatives yield the gradient-magnitude ratios

\begin{equation}
\label{eq:ratio}
\mathrm{Ratio}_x = \left| \frac{\partial \mathcal{L}_{\mathrm{DPO}} / \partial x_w}{\partial \mathcal{L}_{\mathrm{DPO}} / \partial x_l} \right| = \frac{x_l}{x_w},
\qquad
\mathrm{Ratio}_P = \left| \frac{\partial \mathcal{L}_{\mathrm{DPO}} / \partial P_w}{\partial \mathcal{L}_{\mathrm{DPO}} / \partial P_l} \right| = \frac{P_l}{P_w}.
\end{equation}

As training progresses and the policy learns to prefer $y_w$, both ratios fall below 1, meaning DPO suppresses $y_l$ more aggressively than it promotes $y_w$ \citep{feng2024towards}. We note that DPO's gradients with respect to \emph{log-probabilities} are balanced by construction; the asymmetry analyzed by~\citet{feng2024towards} arises in probability and probability-ratio space, where it reflects the differential rate of change of $P_w$ versus $P_l$ during training. AdaDPO operates in this same space, deliberately introducing a compensating asymmetry in log-probability space to maintain promotion pressure on $y_w$ as the policy becomes confident. This imbalance is structural rather than an artifact of hyperparameter choice: equation~\eqref{eq:dpo} applies the same coefficient $\beta$ to both log-ratio terms.

\paragraph{A concrete illustration.}
Consider a single preference pair where, after some training, the policy assigns $P_w = 0.10$ and $P_l = 0.02$ (the model has learned to prefer $y_w$). With a uniform reference $R_w = R_l$, the gradient on $y_w$ is roughly $0.02/0.10 = 1/5$ of the gradient on $y_l$, so the optimization signal on the dispreferred response is five times stronger than on the preferred one. As $P_w$ increases further, the gradient on $y_w$ shrinks toward zero and DPO continues to focus on suppressing $y_l$ rather than reinforcing $y_w$. AdaDPO eliminates this asymmetry by construction (Section~\ref{sec:method}).

\paragraph{Related work.}
DPO variants such as SimPO~\citep{meng2024simpo}, R-DPO~\citep{park2024disentangling}, IPO~\citep{azar2023general}, CPO~\citep{xu2024contrastive}, and ORPO~\citep{hong2024orpo} improve DPO along specific axes yet leave the gradient imbalance unaddressed. While we adopt SimPO's stable parameterization \citep{meng2024simpo}, our work focuses on the orthogonal problem of gradient asymmetry, which persists even in length-normalized settings. A complementary line of work studies \emph{likelihood displacement}~\citep{razin2025unintentional}, where the absolute likelihoods of both responses drift downward, with targeted fixes including DPOP~\citep{pal2024smaug}, Cal-DPO~\citep{xiao2024caldpo}, and SPPO~\citep{wu2024sppo}. Displacement and gradient imbalance are distinct: the former concerns absolute magnitudes of $P_w$ and $P_l$, the latter their relative rate of change within each pair. Crucially, whereas standard DPO is balanced in log-probability space, AdaDPO intentionally introduces logit-space asymmetry to enforce balance in probability space. AdaDPO operates at the per-pair gradient level and is therefore composable with displacement-mitigation methods; its amplification of the gradient on $P_w$ may also partially counteract displacement directly, which we leave for future work.

\paragraph{Concurrent work on gradient imbalance.}
Two concurrent works also target DPO's gradient asymmetry. \citet{ma2025gradient} (Balanced-DPO) introduce a global reweighting factor that balances average gradient contributions across the batch. \citet{zhu2025sgdpo} (SGDPO) add an auxiliary \emph{pilot} term that resamples a sub-sequence from the policy logits. AdaDPO differs by enforcing $|\partial \mathcal{L}/\partial P_w| = |\partial \mathcal{L}/\partial P_l|$ \emph{exactly and per preference pair} via the closed-form ratio of Proposition~\ref{prop:gradient-balance}, with no auxiliary terms, resampling, or extra forward passes, and no new tunable hyperparameters.

\section{AdaDPO: Self-Adaptive Direct Preference Optimization}
\label{sec:method}

\begin{figure}[t]
\centering
\includegraphics[width=\linewidth]{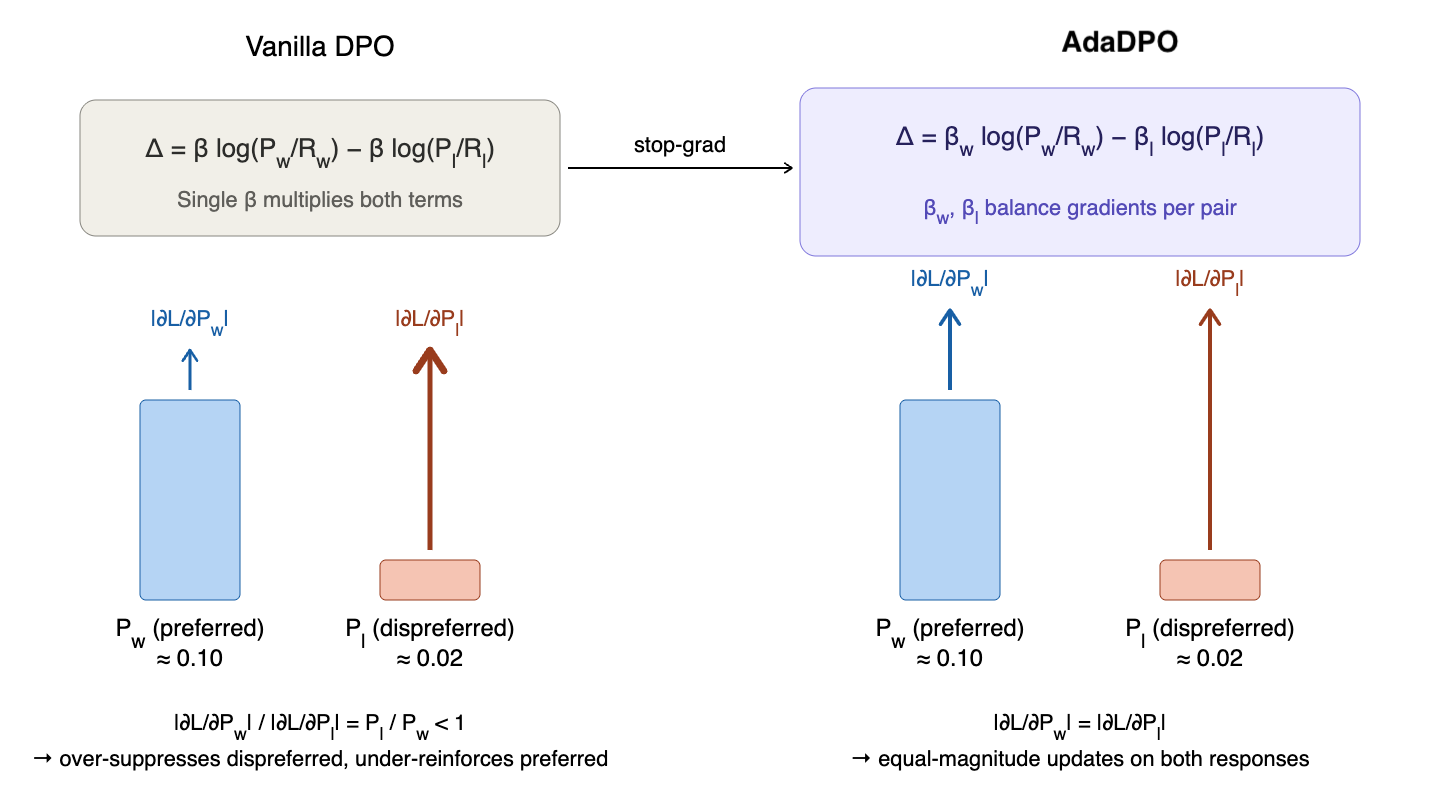}
\caption{Schematic comparison of vanilla DPO (left) and AdaDPO (right). DPO's shared coefficient $\beta$ produces systematically larger gradients on the dispreferred response; AdaDPO's per-pair coefficients enforce equal-magnitude updates on $P_w$ and $P_l$.}
\label{fig:schematic}
\end{figure}

The method follows directly from the gradient analysis: we identify the mathematical condition that produces equal gradient magnitudes on $P_w$ and $P_l$, and construct per-pair coefficients that satisfy this condition exactly. The construction preserves DPO's hyperparameter structure and is implementable in a few lines of code. Figure~\ref{fig:schematic} summarizes the contrast between vanilla DPO and AdaDPO at the per-pair level.

\subsection{Adaptive Coefficients for Balanced Gradients}
\label{sec:method-coefficients}

DPO's implicit reward margin uses the same coefficient $\beta$ for both log-ratio terms~\citep{rafailov2023direct}. We replace it with per-preference-pair coefficients $\beta_w$ and $\beta_l$:
\begin{equation}
\label{eq:adadpo-margin}
\Delta_{\mathrm{AdaDPO}} = \beta_w \log \frac{P_w}{R_w} - \beta_l \log \frac{P_l}{R_l} = r(x, y_w) - r(x, y_l).
\end{equation}

\begin{proposition}[Gradient balance]
\label{prop:gradient-balance}
Let $\mathcal{L}$ be a DPO-style loss with implicit margin $\Delta_{\mathrm{AdaDPO}}$ of the form~\eqref{eq:adadpo-margin}. Requiring $\left|\partial \mathcal{L}/\partial P_w\right| = \left|\partial \mathcal{L}/\partial P_l\right|$ yields the policy-space constraint
\begin{equation}
\label{eq:balance-ratio-policy}
\frac{\beta_w}{\beta_l} = \frac{P_w}{P_l},
\end{equation}
while requiring balance in probability-ratio space, $\left|\partial \mathcal{L}/\partial x_w\right| = \left|\partial \mathcal{L}/\partial x_l\right|$, yields
\begin{equation}
\label{eq:balance-ratio-space}
\frac{\beta_w}{\beta_l} = \frac{x_w}{x_l} = \frac{P_w R_l}{P_l R_w}.
\end{equation}
The two constraints coincide when $R_w = R_l$. In practice, AdaDPO can adopt either balancing rule, yielding strictly equal per-pair gradient magnitudes in the chosen space.
\end{proposition}

\subsection{Stop-Gradient Parameterization}
\label{sec:method-stopgrad}

The ratios in~\eqref{eq:balance-ratio-policy} and~\eqref{eq:balance-ratio-space} depend on $P_w$ and $P_l$, themselves outputs of the policy model. Setting $\beta_w$ directly to these values would propagate gradients through the coefficients and break the intended balance. We treat the coefficients as constants during backpropagation using the stop-gradient operator $\mathrm{sg}(\cdot)$. Setting $\beta_l = \beta$ to preserve DPO's hyperparameter:
\begin{equation}
\label{eq:adadpo-coef}
\beta_w = \beta \cdot \mathrm{sg}\!\left( \frac{P_w}{P_l} \right) \quad \text{or} \quad \beta_w = \beta \cdot \mathrm{sg}\!\left( \frac{P_w R_l}{P_l R_w} \right).
\end{equation}
Both formulations satisfy the balance condition exactly; AdaDPO uses the same $\beta$ values as DPO and adds only a numerical clipping constant $C$ that is robust over a wide range (Appendix~\ref{app:ceiling}), which we view as important for fair comparison and practical adoption. In our experiments we use the ratio-space variant $\beta_w / \beta_l = P_w R_l / (P_l R_w)$, which incorporates the reference policy and corresponds to the implementation in Listing~\ref{lst:adadpo}; an ablation with the policy-space variant $\beta_w / \beta_l = P_w / P_l$ yields essentially identical AlpacaEval~2 performance across the $(\mathrm{lr}, \beta)$ grid, confirming that the choice of balancing space is not critical.

\subsection{The AdaDPO Loss}
\label{sec:method-loss}

The full AdaDPO objective replaces the fixed $\beta$ in DPO's loss with the adaptive coefficients:
\begin{equation}
\label{eq:adadpo-loss}
\mathcal{L}_{\mathrm{AdaDPO}}(\pi_\theta; \pi_{\mathrm{ref}}) = -\mathbb{E}_{(x, y_w, y_l) \sim \mathcal{D}} \left[ \log \sigma\!\left( \beta_w \log \frac{P_w}{R_w} - \beta_l \log \frac{P_l}{R_l} \right) \right].
\end{equation}

Applying Proposition~\ref{prop:gradient-balance} to the AdaDPO loss in~\eqref{eq:adadpo-loss} yields:
\begin{equation}
\label{eq:adadpo-balanced}
\left| \frac{\partial \mathcal{L}_{\mathrm{AdaDPO}} / \partial P_w}{\partial \mathcal{L}_{\mathrm{AdaDPO}} / \partial P_l} \right| = \frac{\beta_w}{\beta_l} \cdot \frac{P_l}{P_w} = 1,
\qquad
\left| \frac{\partial \mathcal{L}_{\mathrm{AdaDPO}} / \partial x_w}{\partial \mathcal{L}_{\mathrm{AdaDPO}} / \partial x_l} \right| = \frac{\beta_w}{\beta_l} \cdot \frac{x_l}{x_w} = \frac{P_w R_l}{P_l R_w} \cdot \frac{P_l R_w}{P_w R_l} = 1.
\end{equation}

By construction, base AdaDPO enforces strict gradient equality between preferred and dispreferred probabilities at every step (Equation~\eqref{eq:adadpo-balanced}), directly correcting the structural asymmetry identified by~\citet{feng2024towards} and rectifying both $\mathrm{Ratio}_x$ and $\mathrm{Ratio}_P$ from Section~\ref{sec:background}. The practical Stable AdaDPO variant used in our experiments approximates this balance per-token within the clipping bound $C$ (Section~\ref{sec:stable-adadpo}).

\paragraph{Intuition for the adaptive ratio.} The coefficients $\beta_w$ and $\beta_l$ are implemented using a stop-gradient operator (e.g., \texttt{detach()} in PyTorch) to ensure they act as constant multipliers during the backward pass rather than differentiable parameters. This choice admits a clear mechanistic interpretation: the gradient on $P_w$ is proportional to $\beta_w / P_w$, while the gradient on $P_l$ is proportional to $\beta_l / P_l$. By setting $\beta_w \propto P_w$, we exactly cancel the inverse-probability scaling that naturally shrinks the gradient as the model becomes confident. This ensures a persistent reinforcement signal that prevents the "vanishing promotion" problem inherent in vanilla DPO, where updates stagnate once the model identifies the preferred response.

\subsection{Implementation}
\label{sec:method-implementation}

AdaDPO requires only a few lines of code on top of a standard DPO loss. We compute the adaptive ratio in log-space (avoiding numerical underflow), clip it at a fixed ceiling $C = 2$ for stability, and use the result to scale $\beta_w$. Listing~\ref{lst:adadpo} shows the complete loss; the modification relative to standard DPO consists of the four lines computing \texttt{beta\_w}.

\begin{lstlisting}[caption={PyTorch implementation of the AdaDPO loss with the length-normalized adaptive ratio used in all main experiments. This corresponds to Stable AdaDPO, formally introduced in Section~\ref{sec:stable-adadpo}; the un-normalized form of Equation~\eqref{eq:adadpo-coef} is recovered by removing the divisions by \texttt{lw} and \texttt{ll}.}, label={lst:adadpo}, basicstyle=\ttfamily\footnotesize, frame=single]
def adadpo_loss(pi_logps, ref_logps, yw_idx, yl_idx, beta,
                C=2.0, yw_lens=None, yl_lens=None):
    pi_yw, pi_yl = pi_logps[yw_idx], pi_logps[yl_idx]
    ref_yw, ref_yl = ref_logps[yw_idx], ref_logps[yl_idx]

    # Adaptive coefficient via stop-gradient (length-normalized)
    lw = yw_lens.float().detach()
    ll = yl_lens.float().detach()
    log_ratio = (pi_yw.detach() / lw - pi_yl.detach() / ll
                 - ref_yw.detach() / lw + ref_yl.detach() / ll)
    beta_w = beta * torch.clamp(log_ratio.exp(), max=C)

    # Standard DPO-style loss with adaptive beta_w
    delta = beta_w * (pi_yw - ref_yw) - beta * (pi_yl - ref_yl)
    return -F.logsigmoid(delta)
\end{lstlisting}

\section{Experimental Setup}
\label{sec:experiments}

\paragraph{Base model and training data.}
Following the Instruct setting of SimPO~\citep{meng2024simpo}, we use Llama-3-8B-Instruct~\citep{aimeta2024llama3} as the base SFT model and train AdaDPO and DPO on the UltraFeedback preference dataset~\citep{cui2024ultrafeedback}, which contains 61{,}837 examples (59{,}876 train, 1{,}961 evaluation). We do not collect any new human data; all experiments use existing open-source preference datasets under their original licenses.

\paragraph{Hyperparameters and grid search.}
We follow SimPO's training settings: batch size 128, maximum sequence length 2048, cosine learning rate schedule with 10\% warmup, training for 1 epoch. We grid-search learning rate $\mathrm{lr} \in \{3, 5, 6, 10\} \times 10^{-7}$ and $\beta \in \{0.005, 0.01, 0.05, 0.1\}$, yielding 16 $(\mathrm{lr}, \beta)$ combinations per method (32 training runs total). Following the standard reporting practice in this literature, we evaluate every grid cell on each benchmark and report both the best cell (Table~\ref{tab:main}) and aggregate statistics across the full hyperparameter grid (Section~\ref{sec:results-sweep}); the latter is robust to the choice of best-cell reporting convention.

\paragraph{Evaluation benchmarks.}
We evaluate on three standard instruction-following benchmarks with GPT-4 Turbo as the judge. As is standard in LLM alignment work, we rely on GPT-4-Turbo as an automatic evaluator and therefore inherit its known biases (e.g., verbosity preference and occasional inconsistency), which we partially control for via length-controlled metrics but do not eliminate entirely. \textbf{AlpacaEval 2}~\citep{dubois2024length} consists of 805 queries and reports length-controlled win rate (LC), raw win rate (WR), and average response length. \textbf{Arena-Hard v0.1}~\citep{li2024arenahard} contains 500 challenging technical queries and reports WR with 95\% confidence intervals. \textbf{MT-Bench}~\citep{zheng2023judging} covers 8 categories with 80 multi-turn questions on a 1--10 scale.

\section{Results}
\label{sec:results}

Throughout this section, ``AdaDPO'' refers to the length-normalized variant (Stable AdaDPO; Section~\ref{sec:stable-adadpo}), which is the implementation shown in Listing~\ref{lst:adadpo}. The base form of Equation~\eqref{eq:adadpo-coef} yields comparable best-case performance after tuning but exhibits less stable training dynamics on long responses (Appendix~\ref{app:implementation}).

\subsection{Optimization Dynamics}
\label{sec:results-dynamics}

\begin{figure}[t]
  \centering
  \includegraphics[width=\linewidth]{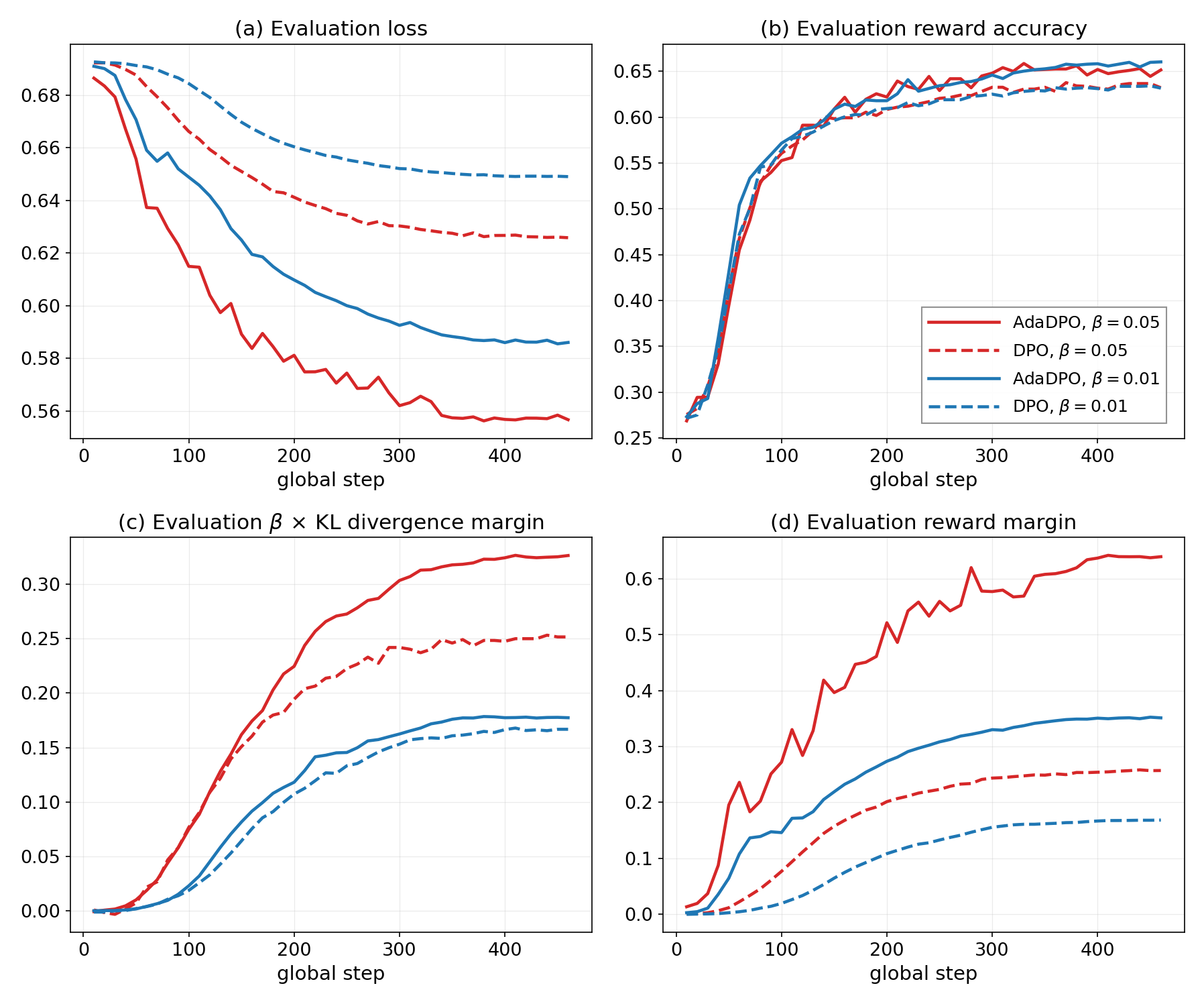}
  \caption{Training-dynamics comparison on the validation set. (a) Evaluation loss; (b) reward accuracy ($\Pr[r(x, y_w) > r(x, y_l)]$); (c) $\beta \times \mathrm{KL}$ divergence margin; (d) reward margin $r(x, y_w) - r(x, y_l)$. Solid: AdaDPO, dashed: DPO; red: $\beta = 0.05$, blue: $\beta = 0.01$. AdaDPO's loss and reward margin are scaled by the adaptive coefficient $\beta_w$, so absolute values are not directly comparable to DPO's; reward accuracy~(b) is scale-invariant.}
  \label{fig:dynamics}
\end{figure}

Figure~\ref{fig:dynamics} compares AdaDPO and DPO under two representative $\beta$ values (0.05 and 0.01) at fixed learning rate $5 \times 10^{-7}$. AdaDPO consistently achieves lower evaluation loss (Figure~\ref{fig:dynamics}a), higher reward accuracy (Figure~\ref{fig:dynamics}b), larger $\beta \times \mathrm{KL}$ divergence margin (Figure~\ref{fig:dynamics}c), and larger reward margin $r(x, y_w) - r(x, y_l)$ (Figure~\ref{fig:dynamics}d). The performance ordering across all metrics consistently places AdaDPO above DPO at matched $\beta$. While the $\beta \times \mathrm{KL}$ margin is dominated by the KL constraint strength, the other metrics---more directly tied to alignment quality---benefit clearly from balanced gradient updates, indicating that a large KL margin alone is insufficient: balanced updates are key to translating margin into improved performance.

\subsection{Best Performance Comparison}
\label{sec:results-best}

Table~\ref{tab:main} reports the strongest configuration per method across the 16 $(\mathrm{lr}, \beta)$ combinations, while Section~\ref{sec:results-sweep} analyzes the full hyperparameter grid. AdaDPO achieves the global best LC (48.3\%) and WR (46.1\%) on AlpacaEval 2, with shorter average response length than DPO on that benchmark. On Arena-Hard (Figure~\ref{fig:arenahard} in Appendix~\ref{app:additional}), DPO achieves a slightly higher WR (42.9\% vs.\ 41.5\%), but the 95\% confidence intervals overlap substantially ([41.1, 44.6] vs.\ [39.7, 43.1]), indicating no statistically significant difference between the two methods. Average response lengths on Arena-Hard are similar (DPO: 525, AdaDPO: 530), and the benchmark does not provide a length-controlled metric, so we cannot fully separate length effects from response quality on this benchmark. On MT-Bench (Figure~\ref{fig:mtbench} in Appendix~\ref{app:additional}), AdaDPO (8.03) marginally exceeds DPO (8.01); the 0.02-point gap is below the resolution of MT-Bench's 1--10 scale and 80-question evaluation set, so we do not interpret this benchmark as differentiating the methods.

\begin{table}[b]
\caption{Best performance per method across the 16 $(\mathrm{lr}, \beta)$ configurations. AdaDPO achieves the global best LC and WR on AlpacaEval~2 with the shortest average length on that benchmark.}
\label{tab:main}
\centering
\small
\begin{tabular}{lccccccccc}
\toprule
& \multicolumn{4}{c}{AlpacaEval 2} & \multicolumn{4}{c}{Arena-Hard} & MT-Bench \\
\cmidrule(lr){2-5} \cmidrule(lr){6-9} \cmidrule(lr){10-10}
Method & LC (\%) & WR (\%) & STD & Length & WR (\%) & 95\%-Hi & 95\%-Lo & Length & Score \\
\midrule
SFT      & 28.5 & 28.8 & 1.6 & 1976 & 26.3 & 28.1 & 24.7 & 589 & 7.58 \\
DPO      & 46.4 & 44.5 & 1.8 & 1933 & \textbf{42.9} & 44.6 & 41.1 & 525 & 8.01 \\
AdaDPO   & \textbf{48.3} & \textbf{46.1} & 1.8 & 1908 & 41.5 & 43.1 & 39.7 & 530 & \textbf{8.03} \\
\bottomrule
\end{tabular}
\end{table}

\subsection{Hyperparameter Sweep on AlpacaEval 2}
\label{sec:results-sweep}

\begin{figure}[ht]
  \centering
  \includegraphics[width=\linewidth]{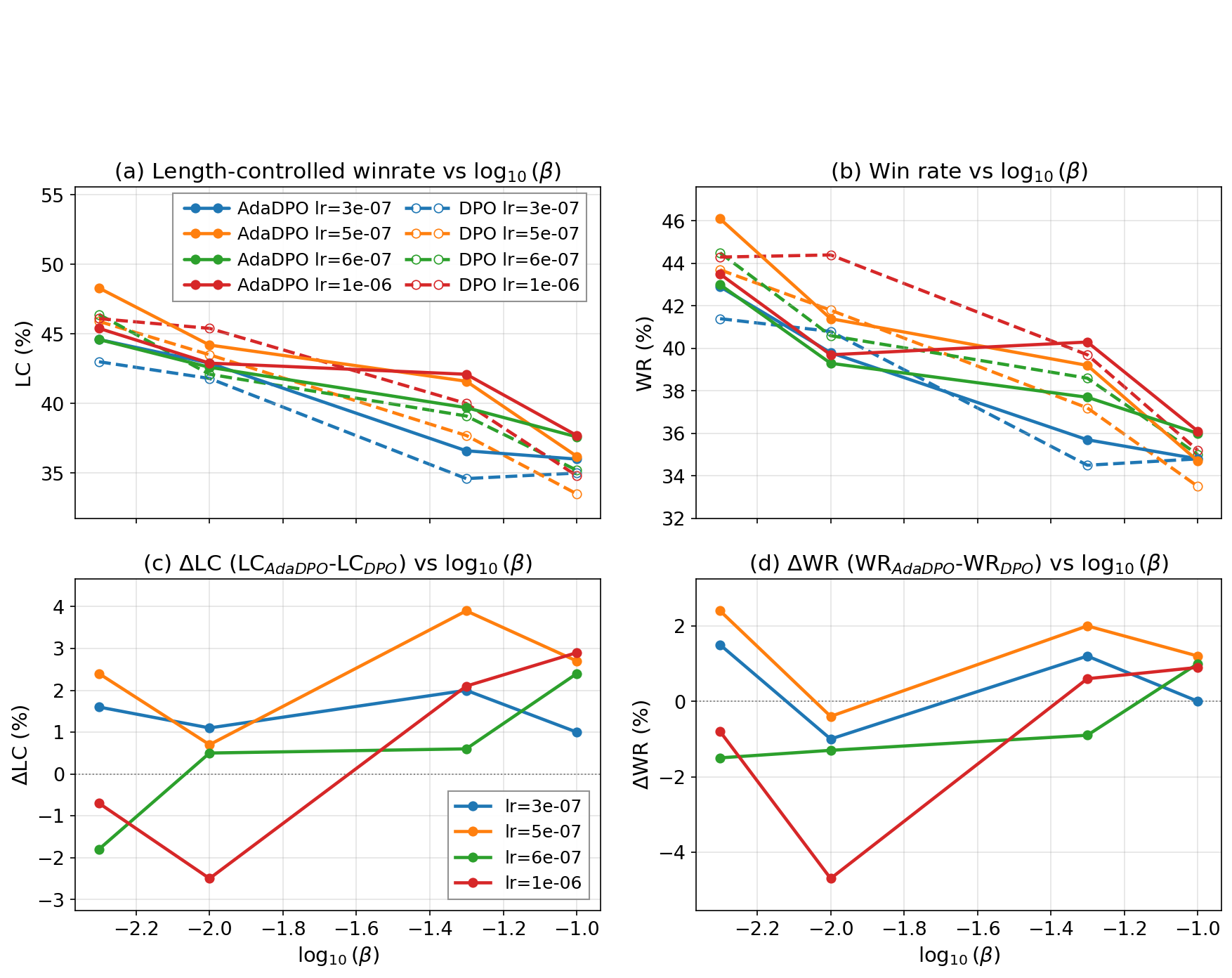}
 \caption{AlpacaEval~2 performance vs.\ $\log_{10}(\beta)$. (a) LC; (b) WR; (c) $\Delta\mathrm{LC} = \mathrm{LC}_{\mathrm{AdaDPO}} - \mathrm{LC}_{\mathrm{DPO}}$; (d) $\Delta\mathrm{WR}$. Solid: AdaDPO; dashed: DPO.}
  \label{fig:alpacaeval}
\end{figure}

To rigorously compare AdaDPO and DPO, we evaluate each of the 16 $(\mathrm{lr}, \beta)$ combinations on AlpacaEval 2, rather than relying on a single tuned configuration. Figure~\ref{fig:alpacaeval} visualizes the results.

Aggregating across the 16 combinations:
\begin{itemize}
\item \textbf{LC}: AdaDPO outperforms DPO in 13/16 (81\%) combinations.
\item \textbf{WR}: AdaDPO outperforms DPO in 9/16 (56\%) combinations.
\item \textbf{$\Delta = \mathrm{LC} - \mathrm{WR}$}: AdaDPO produces a larger LC-over-WR margin in 14/16 (88\%) combinations, indicating reduced length exploitation.
\end{itemize}

Furthermore, 31 out of 32 runs (97\%) satisfy LC $\geq$ WR. These results confirm that AdaDPO consistently improves length-controlled performance across the majority of hyperparameter settings while effectively mitigating length bias.

\paragraph{Length-bias mitigation in detail.} 
The aggregate $\mathrm{LC}-\mathrm{WR}$ margin gain (88\% of combinations) is the most direct evidence that AdaDPO reduces length exploitation, but it can be confirmed at finer granularity. Across the 16 hyperparameter combinations, AdaDPO produces shorter average responses than DPO in 12/16 (75\%) configurations on the AlpacaEval 2 benchmark, while simultaneously achieving higher LC, indicating that AdaDPO's quality gains are not driven by verbosity. The strongest LC improvements ($\Delta \mathrm{LC} \geq 2$pp) occur at the shortest average response lengths, consistent with the theoretical prediction that balanced gradients reduce the implicit incentive for length exploitation present in DPO~\citep{park2024disentangling}.

\section{Generalization to Other DPO Variants}
\label{sec:generalization}

The AdaDPO principle---replacing fixed coefficients with self-adaptive stop-gradient-based coefficients---applies to any pairwise contrastive loss of the form
\begin{equation}
\mathcal{L} = f\!\left( \beta_w \log \frac{P_w}{R_w} - \beta_l \log \frac{P_l}{R_l} + \lambda \right),
\end{equation}
where $f$ is any differentiable function (e.g., $f(x) = \log \sigma(x)$ for DPO, SimPO, R-DPO, CPO; $f(x) = x^2$ for IPO). Imposing the gradient-balance condition yields $\beta_w / \beta_l = \mathrm{sg}(P_w / P_l)$ or $\mathrm{sg}(P_w R_l / P_l R_w)$. The principle extends to ORPO with a slight modification: $\beta_w / \beta_l = \mathrm{sg}\!\left(P_w(1 - P_w) / [P_l(1 - P_l)]\right)$. Table~\ref{tab:adadpo-variants} summarizes the resulting self-adaptive variants for each method, preserving the original hyperparameters of each base method. Detailed derivations are presented in Appendix~\ref{app:generalization}.

\subsection{Stable AdaDPO: A Length-Normalized Variant}
\label{sec:stable-adadpo}

Computing the adaptive ratio over token-summed log-probabilities can produce unstable training dynamics on long responses, where small per-token differences accumulate into large ratios that frequently saturate the clipping bound $C$. Inspired by SimPO's length-normalized reward parameterization, \emph{Stable AdaDPO} computes the adaptive ratio over per-token average log-probabilities. Formally, this corresponds to setting exponents $\alpha_w = 1/|y_w|$, $\alpha_l = 1/|y_l|$ in the generalization
\begin{equation}
\label{eq:stable-adadpo}
\beta_w = \beta \cdot \mathrm{sg}\!\left( \frac{(P_w/R_w)^{\alpha_w}}{(P_l/R_l)^{\alpha_l}} \right),
\end{equation}
which recovers DPO when $\alpha_w = \alpha_l = 0$ and base AdaDPO when $\alpha_w = \alpha_l = 1$. As noted at the start of Section~\ref{sec:results}, this is the variant used in all main experiments: length normalization yields smoother log-probability curves throughout training and more robust convergence than the unnormalized formulation, particularly at small $\beta$ where gradient magnitudes are sensitive to the exact ratio. The full $(\alpha_w, \alpha_l)$ family is discussed in Appendix~\ref{app:stable}.

\begin{table}[t]
\centering
\caption{Self-adaptive counterparts of preference-optimization objectives under the AdaDPO principle. Each method preserves its original hyperparameters (DPO, CPO, R-DPO, ORPO, and SimPO use $\beta$; IPO has no $\beta$, hence $\beta_l = 1$).}
\label{tab:adadpo-variants}
\begin{tabular}{l l l}
\toprule
Method & Self-adaptive objective & Adaptive coefficients \\
\midrule
DPO &
$\log \sigma\!\big(\beta_w \log \tfrac{P_w}{R_w} - \beta_l \log \tfrac{P_l}{R_l}\big)$ &
$\beta_l = \beta,\;\; \beta_w = \beta \cdot \mathrm{sg}\!\big(\tfrac{P_w R_l}{P_l R_w}\big)$ \\
\addlinespace[0.4em]
IPO &
$\big(\beta_w \log \tfrac{P_w}{R_w} - \beta_l \log \tfrac{P_l}{R_l} - \tfrac{1}{2\tau}\big)^{\!2}$ &
$\beta_l = 1,\;\; \beta_w = \mathrm{sg}\!\big(\tfrac{P_w R_l}{P_l R_w}\big)$ \\
\addlinespace[0.4em]
R-DPO &
$\log \sigma\!\big(\beta_w \log \tfrac{P_w}{R_w} - \beta_l \log \tfrac{P_l}{R_l} - \alpha(|y_w| - |y_l|)\big)$ &
$\beta_l = \beta,\;\; \beta_w = \beta \cdot \mathrm{sg}\!\big(\tfrac{P_w R_l}{P_l R_w}\big)$ \\
\addlinespace[0.4em]
CPO &
$\log \sigma\!\big(\beta_w \log P_w - \beta_l \log P_l\big)$ &
$\beta_l = \beta,\;\; \beta_w = \beta \cdot \mathrm{sg}\!\big(\tfrac{P_w}{P_l}\big)$ \\
\addlinespace[0.4em]
ORPO &
$\log P_w + \log \sigma\!\big(\beta_w \log \tfrac{P_w}{1 - P_w} - \beta_l \log \tfrac{P_l}{1 - P_l}\big)$ &
$\beta_l = \beta,\;\; \beta_w = \beta \cdot \mathrm{sg}\!\big(\tfrac{P_w (1 - P_w)}{P_l (1 - P_l)}\big)$ \\
\addlinespace[0.4em]
SimPO &
$\log \sigma\!\big(\tfrac{\beta_w}{|y_w|} \log P_w - \tfrac{\beta_l}{|y_l|} \log P_l - \gamma\big)$ &
$\beta_l = \beta,\;\; \beta_w = \beta \cdot \mathrm{sg}\!\big(\tfrac{|y_w| P_w}{|y_l| P_l}\big)$ \\
\bottomrule
\end{tabular}
\end{table}

\section{Discussion and Limitations}
\label{sec:discussion}

\paragraph{Why AdaDPO works.}
AdaDPO counteracts the asymmetric gradient ratio of equation~\eqref{eq:ratio} by setting $\beta_w / \beta_l$ to scale with $P_w / P_l$ (or $P_w R_l / P_l R_w$): when the policy becomes much more likely to generate $y_w$ than $y_l$, the gradient on $y_w$ would otherwise vanish, and AdaDPO amplifies it to match the gradient on $y_l$. Empirically, this balance translates into observable gains in reward accuracy, reward margin, and length-controlled win rate.

\paragraph{Stability of the Ceiling Constant.} 
In AdaDPO, the adaptive coefficient $\beta_w$ is clipped by a ceiling constant $C$ to ensure numerical stability. While $C=2$ was used for all main experiments, our ablation study in Appendix~\ref{app:ceiling} confirms that performance is robust across the range $C \in [1.5, 2.5]$. Training only becomes unstable when $C$ exceeds 5, at which point the adaptive coefficients can amplify gradients enough to cause instability. This relative insensitivity suggests that $C$ is a stable architectural choice rather than a hyperparameter requiring intensive per-task tuning.

\paragraph{Limitations.}
\textbf{(i) Single base model.} We use Llama-3-8B-Instruct; cross-model validation on Mistral, Qwen, and larger Llama checkpoints remains future work.
\textbf{(ii) Single preference dataset.} We train on UltraFeedback only; transfer to other preference-data sources (e.g., HH-RLHF, PKU-SafeRLHF) is untested.
\textbf{(iii) Arena-Hard underperformance.} On Arena-Hard, DPO marginally outperforms AdaDPO in raw win rate; the 95\% confidence intervals overlap substantially, so the gap is not statistically significant. Arena-Hard does not provide a length-controlled metric, limiting our ability to disentangle length effects from quality on this benchmark.
\textbf{(iv) Scope of claims.} Our results apply to offline preference optimization in the LLM Instruct setting; extending AdaDPO to online RL, other modalities, and stronger safety objectives remains future work rather than a claim of this paper.

\section{Conclusion}
\label{sec:conclusion}

We identified the gradient imbalance in DPO as a structural property of its loss that causes models to suppress dispreferred responses far more aggressively than they promote preferred ones. AdaDPO resolves this by introducing per-preference-pair adaptive coefficients that enforce equality of gradient magnitudes on $P_w$ and $P_l$ (exactly in the base form, and per-token within a clipping bound in Stable AdaDPO implementation), with minimal code changes and no sensitive hyperparameters beyond DPO's $\beta$. Two observations follow from our results. First, the consistency of AdaDPO's gains across the 16-cell hyperparameter grid---LC win in 81\% of combinations and an enlarged LC-over-WR margin in 88\%---suggests that gradient balance, rather than careful $\beta$ tuning, is the constraint on length-controlled performance for DPO-style objectives. Second, the LC-over-WR margin improvement systematically tracks reduced response lengths, providing direct empirical evidence that DPO's gradient asymmetry is mechanistically linked to length exploitation, and offering an alternative to the explicit length regularization used in prior work. Cross-model validation, transfer to alternative preference datasets, and empirical evaluation of the variants in Table~\ref{tab:adadpo-variants} are natural next steps. More broadly, the per-pair gradient-balance condition appears to be a useful design lens for preference optimization beyond any specific loss function.

\begin{ack}
The authors would like to thank Lei Li and Nazhou Liu for their insightful suggestions, as well as Chenghua Wang, Zhifeng Li, Wenhui Chen, Jie Zhou, Xinmiao Yu, and Yanlin Fei for their valuable feedback and discussions. 
\end{ack}

\clearpage
\bibliographystyle{plainnat}
\bibliography{references}

@inproceedings{christiano2017deep,
  title={Deep reinforcement learning from human preferences},
  author={Christiano, Paul F and Leike, Jan and Brown, Tom and Martic, Miljan and Legg, Shane and Amodei, Dario},
  booktitle={Advances in Neural Information Processing Systems (NeurIPS)},
  year={2017}
}

@article{ziegler2019fine,
  title={Fine-tuning language models from human preferences},
  author={Ziegler, Daniel M and Stiennon, Nisan and Wu, Jeffrey and Brown, Tom B and Radford, Alec and Amodei, Dario and Christiano, Paul and Irving, Geoffrey},
  journal={arXiv preprint arXiv:1909.08593},
  year={2019}
}

@article{schulman2017proximal,
  title={Proximal policy optimization algorithms},
  author={Schulman, John and Wolski, Filip and Dhariwal, Prafulla and Radford, Alec and Klimov, Oleg},
  journal={arXiv preprint arXiv:1707.06347},
  year={2017}
}

@inproceedings{rafailov2023direct,
  title={Direct preference optimization: Your language model is secretly a reward model},
  author={Rafailov, Rafael and Sharma, Archit and Mitchell, Eric and Ermon, Stefano and Manning, Christopher D and Finn, Chelsea},
  booktitle={Advances in Neural Information Processing Systems (NeurIPS)},
  year={2023}
}

@article{feng2024towards,
  title={Towards analyzing and understanding the limitations of {DPO}: A theoretical perspective},
  author={Feng, Duanyu and Qin, Bowen and Huang, Chen and Zhang, Zheng and Lei, Wenqiang},
  journal={arXiv preprint arXiv:2404.04626},
  year={2024}
}

@article{meng2024simpo,
  title={{SimPO}: Simple preference optimization with a reference-free reward},
  author={Meng, Yu and Xia, Mengzhou and Chen, Danqi},
  journal={arXiv preprint arXiv:2405.14734},
  year={2024}
}

@inproceedings{cui2024ultrafeedback,
  title={{UltraFeedback}: Boosting language models with high-quality feedback},
  author={Cui, Ganqu and Yuan, Lifan and Ding, Ning and Yao, Guanming and Zhu, Wei and Ni, Yuan and Xie, Guotong and Liu, Zhiyuan and Sun, Maosong},
  booktitle={International Conference on Machine Learning (ICML)},
  year={2024}
}

@article{dubois2024length,
  title={Length-controlled {AlpacaEval}: A simple way to debias automatic evaluators},
  author={Dubois, Yann and Galambosi, Bal{\'a}zs and Liang, Percy and Hashimoto, Tatsunori B},
  journal={arXiv preprint arXiv:2404.04475},
  year={2024}
}

@article{park2024disentangling,
  title={Disentangling length from quality in direct preference optimization},
  author={Park, Ryan and Rafailov, Rafael and Ermon, Stefano and Finn, Chelsea},
  journal={arXiv preprint arXiv:2403.19159},
  year={2024}
}

@article{azar2023general,
  title={A general theoretical paradigm to understand learning from human preferences},
  author={Azar, Mohammad Gheshlaghi and Rowland, Mark and Piot, Bilal and Guo, Daniel and Calandriello, Daniele and Valko, Michal and Munos, R{\'e}mi},
  journal={arXiv preprint arXiv:2310.12036},
  year={2023}
}

@article{xu2024contrastive,
  title={Contrastive preference optimization: Pushing the boundaries of {LLM} performance in machine translation},
  author={Xu, Haoran and Sharaf, Amr and Chen, Yunmo and Tan, Weiting and Shen, Lingfeng and Van Durme, Benjamin and Murray, Kenton and Kim, Young Jin},
  journal={arXiv preprint arXiv:2401.08417},
  year={2024}
}

@article{hong2024orpo,
  title={{ORPO}: Monolithic preference optimization without reference model},
  author={Hong, Jiwoo and Lee, Noah and Thorne, James},
  journal={arXiv preprint arXiv:2403.07691},
  year={2024}
}

@misc{li2024arenahard,
  author = {Tianle Li and Wei-Lin Chiang and Evan Frick and Lisa Dunlap and Tianhao Wu and Banghua Zhu and Joseph E. Gonzalez and Ion Stoica},
  title  = {From Live Data to High-Quality Benchmarks: The {Arena-Hard} Pipeline},
  year   = {2024},
  howpublished = {LMSYS Blog},
  note   = {\url{https://lmsys.org/blog/2024-04-19-arena-hard/}}
}

@inproceedings{zheng2023judging,
  title={Judging {LLM-as-a-judge} with {MT-Bench} and {Chatbot Arena}},
  author={Zheng, Lianmin and Chiang, Wei-Lin and Sheng, Ying and Zhuang, Siyuan and Wu, Zhanghao and Zhuang, Yonghao and Lin, Zi and Li, Zhuohan and Li, Dacheng and Xing, Eric P and Zhang, Hao and Gonzalez, Joseph E and Stoica, Ion},
  booktitle={Advances in Neural Information Processing Systems (NeurIPS) Datasets and Benchmarks Track},
  year={2023}
}

@article{aimeta2024llama3,
  title={The {Llama 3} herd of models},
  author={{AI@Meta}},
  journal={arXiv preprint arXiv:2407.21783},
  year={2024}
}

@inproceedings{razin2025unintentional,
  title={Unintentional unalignment: Likelihood displacement in direct preference optimization},
  author={Razin, Noam and Malladi, Sadhika and Bhaskar, Adithya and Chen, Danqi and Arora, Sanjeev and Hanin, Boris},
  booktitle={International Conference on Learning Representations (ICLR)},
  year={2025}
}

@article{pal2024smaug,
  title={Smaug: Fixing failure modes of preference optimisation with {DPO}-positive},
  author={Pal, Arka and Karkhanis, Deep and Dooley, Samuel and Roberts, Manley and Naidu, Siddartha and White, Colin},
  journal={arXiv preprint arXiv:2402.13228},
  year={2024}
}

@inproceedings{xiao2024caldpo,
  title={{Cal-DPO}: Calibrated direct preference optimization for language model alignment},
  author={Xiao, Teng and Yuan, Yige and Zhu, Huaisheng and Li, Mingxiao and Honavar, Vasant G},
  booktitle={Advances in Neural Information Processing Systems (NeurIPS)},
  year={2024}
}

@article{wu2024sppo,
  title={Self-play preference optimization for language model alignment},
  author={Wu, Yue and Sun, Zhiqing and Yuan, Huizhuo and Ji, Kaixuan and Yang, Yiming and Gu, Quanquan},
  journal={arXiv preprint arXiv:2405.00675},
  year={2024}
}

@article{ma2025gradient,
  author  = {Qinwei Ma and Jingzhe Shi and Can Jin and Jenq-Neng Hwang and Serge Belongie and Lei Li},
  title   = {Gradient Imbalance in Direct Preference Optimization},
  journal = {arXiv preprint arXiv:2502.20847},
  year    = {2025}
}

@inproceedings{zhu2025sgdpo,
  author    = {Wenqiao Zhu and Ji Liu and Lulu Wang and Jun Wu and Yulun Zhang},
  title     = {{SGDPO}: Self-Guided Direct Preference Optimization for Language Model Alignment},
  booktitle = {Findings of the Association for Computational Linguistics: ACL 2025},
  pages     = {12366--12383},
  year      = {2025},
  publisher = {Association for Computational Linguistics},
  address   = {Vienna, Austria}
}


\clearpage
\appendix

\section{Technical appendices and supplementary material}

\subsection{Implementation Details}
\label{app:implementation}

The complete PyTorch implementation appears in Listing~\ref{lst:adadpo} of Section~\ref{sec:method-implementation}. We use the length-normalized variant (Stable AdaDPO; see Section~\ref{sec:stable-adadpo}) for all main-text experiments. The unnormalized variant---obtained by replacing the per-token-averaged log-probabilities with raw summed log-probabilities---produces less stable training dynamics on long responses but achieves comparable best-case performance after appropriate hyperparameter tuning. The full grid-search artifacts (training logs, evaluation outputs, and per-configuration metrics for all 32 runs) will be released with the camera-ready version.

\subsection{Ceiling Constant Ablation}
\label{app:ceiling}

We ablate the ceiling constant $C$ over $\{1.5, 2, 2.5, 3, 4, 5, 10\}$ at $\beta = 0.01$, taking the best LC across two learning rates ($5 \times 10^{-7}$ and $1 \times 10^{-6}$) per value of $C$. The optimum is $C = 2$ (LC 44.2\%, WR 42.7\%); performance is robust on $[1.5, 2.5]$ and degrades beyond $C \geq 5$.

\begin{table}[h]
\caption{Effect of ceiling constant $C$ on AlpacaEval 2 performance ($\beta = 0.01$).}
\label{tab:ceiling}
\centering
\small
\begin{tabular}{cccc}
\toprule
$C$ & LC (\%) & WR (\%) & Length \\
\midrule
1.5 & 43.7 & 37.3 & 1772 \\
2.0 & \textbf{44.2} & \textbf{42.7} & 1941 \\
2.5 & 40.5 & 37.5 & 1867 \\
3.0 & 41.6 & 37.8 & 1838 \\
4.0 & 39.8 & 37.1 & 1881 \\
5.0 & 38.4 & 36.0 & 1886 \\
10.0 & 34.9 & 34.8 & 1979 \\
\bottomrule
\end{tabular}
\end{table}

\subsection{Generalization to DPO Variants: Derivations}
\label{app:generalization}

For any pairwise contrastive loss $\mathcal{L} = f(\beta_w \log P_w - \beta_l \log P_l + \lambda)$ or $\mathcal{L} = f(\beta_w \log x_w - \beta_l \log x_l + \lambda)$, requiring $|\partial \mathcal{L}/\partial P_w| = |\partial \mathcal{L}/\partial P_l|$ yields $\beta_w / \beta_l = P_w / P_l$ (or equivalently $\beta_w / \beta_l = P_w R_l / (P_l R_w)$).

Beyond symmetric balancing ($k = 1$), the same framework allows an optional scaling factor $k$ that biases the update toward reinforcing preferred responses ($k > 1$) or suppressing dispreferred responses ($k < 1$). Concretely, one can set
\[
\frac{\beta_w}{\beta_l} = k \cdot \mathrm{sg}\!\left(\frac{P_w}{P_l}\right),
\]
leaving all other aspects of the loss unchanged. In this paper we always fix $k = 1$ and report results only for the balanced setting, which we find to be the most robust across general benchmarks.

Using the stop-gradient operator gives the per-method coefficients summarized in Table~\ref{tab:adadpo-variants}. For ORPO, whose loss involves $\log[P/(1-P)]$, the balance condition gives $\beta_w / \beta_l = \mathrm{sg}\!\left(P_w(1-P_w)/[P_l(1-P_l)]\right)$. The exponent generalization framework discussed in Section~\ref{sec:stable-adadpo} extends naturally to all variants in Table~\ref{tab:adadpo-variants} by replacing each adaptive coefficient with its $(\alpha_w, \alpha_l)$-parameterized counterpart.

\subsection{Stable AdaDPO: Additional Details}
\label{app:stable}

The exponent family introduced in equation~\eqref{eq:stable-adadpo} accommodates several special cases beyond DPO ($\alpha = 0$), base AdaDPO ($\alpha = 1$), and Stable AdaDPO ($\alpha = 1/|y|$). Intermediate values of $\alpha$ smoothly interpolate between these regimes. We empirically observed that Stable AdaDPO yields the most stable training across the broadest range of $\beta$ values; hence its use throughout Section~\ref{sec:results}. Asymmetric exponents $(\alpha_w \neq \alpha_l)$ are an interesting direction for future work, particularly for settings where preferred and dispreferred responses differ systematically in length or complexity (e.g., reasoning-style preferences with verbose chain-of-thought $y_w$ and terse incorrect $y_l$).

\subsection{Additional Benchmark Results}
\label{app:additional}

\begin{figure}[h]
  \centering
  \includegraphics[width=\linewidth]{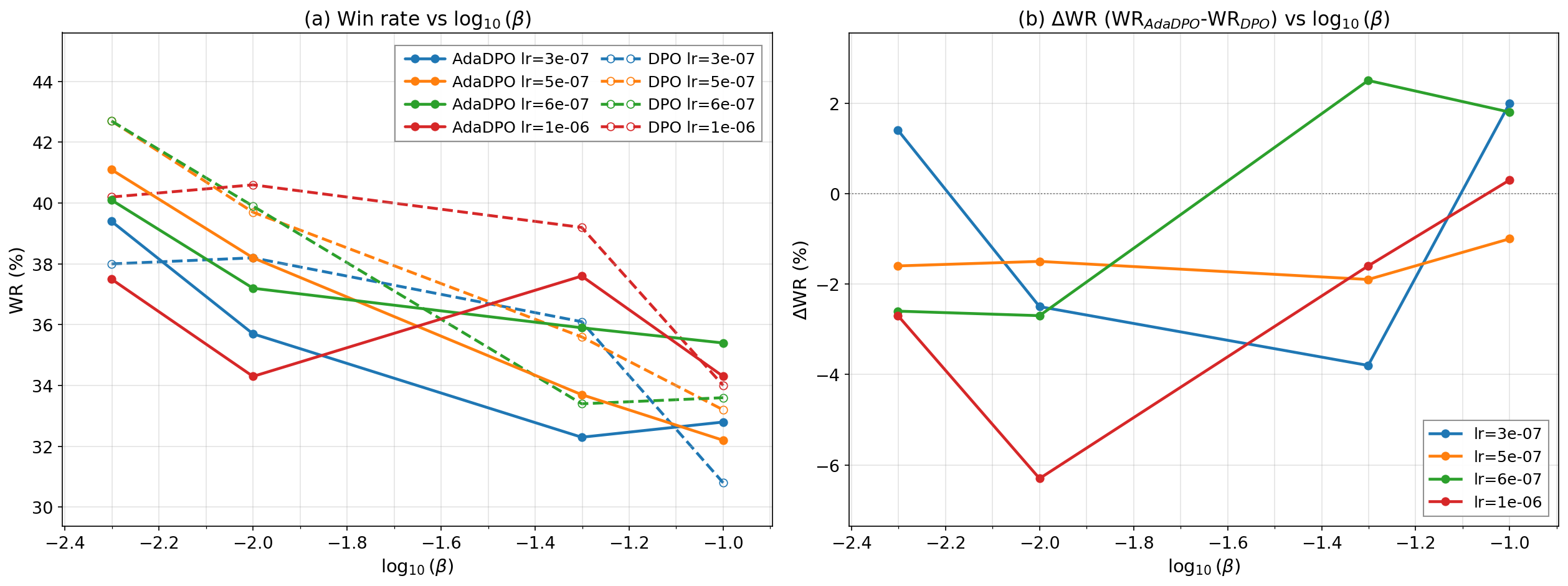}
  \caption{Arena-Hard performance comparison between AdaDPO and DPO. (a) Win rate vs.\ $\beta$; (b) $\Delta\mathrm{WR}$. The 95\% confidence intervals on the best configurations overlap, indicating no statistically significant difference between methods on this length-uncontrolled benchmark.}
  \label{fig:arenahard}
\end{figure}

\begin{figure}[h]
  \centering
  \includegraphics[width=\linewidth]{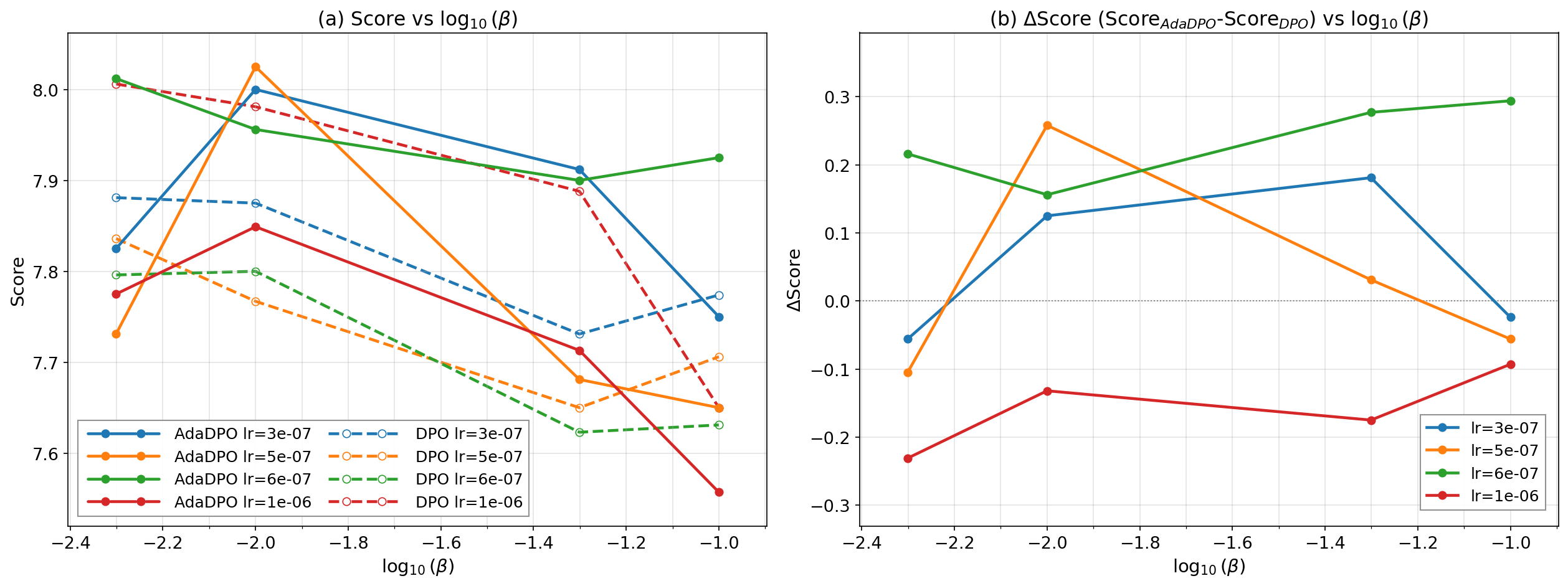}
  \caption{MT-Bench performance comparison between AdaDPO and DPO. The 0.02-point gap between AdaDPO (8.03) and DPO (8.01) is below the resolution of the 1--10 scoring scale and 80-question evaluation set.}
  \label{fig:mtbench}
\end{figure}

\end{document}